\documentclass[11pt]{article}

\usepackage{agenticlearning-xelatex}

\usepackage{caption}

\providecommand{\FourAFCfigFrac}{0.46}%
\providecommand{\FourAFCtabFrac}{0.52}%
\providecommand{\MmPolicyLeftFrac}{0.52}%
\providecommand{\MmPolicyRightFrac}{0.44}%
\providecommand{\ProbeVtwtLeftFrac}{0.50}%
\providecommand{\ProbeVtwtRightFrac}{0.46}%

\begin{document}

\shorttitle{Continual Visual and Verbal Learning}
\shortauthor{Xiaoyang Jiang \etal}

\title{Continual Visual and Verbal Learning \\
Through a Child's Egocentric Input}
\author{%
  Xiaoyang Jiang$^{1}$,
  Yanlai Yang$^{1}$,
  Kenneth A.~Norman$^{2}$,
  Brenden Lake$^{2}$,
  Mengye Ren$^{1}$ \\
  $^{1}$Agentic Learning AI Lab, New York University \\
  $^{2}$Department of Psychology, Princeton University\\
  \texttt{\{xj2366, yy2694, mengye\}@nyu.edu}
}
\date{}
\maketitle

\begin{abstract}
Children learn the meanings of words from a continuous, temporally structured stream of egocentric experience. Recent work shows that neural networks can also learn word-referent mappings from a child's egocentric video recordings, but they cycle through the shuffled data for hundreds of epochs, contrasting with how children actually encounter their environment. We introduce BabyCL, a continual multimodal learning framework that processes the SAYCam dataset in a single chronological pass, combining streaming visual representation learning with an image-text contrastive objective. BabyCL combines a multi-stage temporal segmentation of the stream with a dual replay buffer that independently manages visual and multimodal histories, and it is jointly trained with three contrastive losses on a shared backbone. Under a matched optimization budget, BabyCL outperforms streaming learning baselines on the SAYCam Labeled-S 4AFC benchmark, substantially narrowing the gap to an upper bound of offline training. Ablations show that the gains are robust to the length of the online temporal segmentation window and the eviction rule of the replay buffer. Together, these results show that meaningful word-referent mappings can emerge under training conditions much closer to a child's actual experience.
\end{abstract}

\section{Introduction}
Well before their first birthday, children begin learning their first words, linking spoken words to their visual referents~\citep{bergelson2012months}. By age two, children are highly proficient word learners~\citep{bloom2002children}: they can associate spoken words with objects, actions, and scenes that they encounter in everyday life. Children learn language despite a continuous stream of input that is noisy, temporally correlated, and highly redundant. Whether this natural input stream is sufficient for word learning or we need strong linguistic priors is a central question at the intersection of cognitive science and artificial intelligence.

One mechanism that could underlie this ability is \emph{cross-situational word learning}, whereby children (or computational models) are sensitive to cross-situational regularities between visual and verbal input~\citep{yu2007rapid, goodman2007bayesian}. For example,~\citet{vong2022cross} showed that multimodal neural networks can also acquire word-referent mappings through cross-situational statistics, but it (and related models) rely on curated laboratory stimuli or small-scale datasets. It is unclear whether such mechanisms scale to noisy, continuous, and temporally structured inputs that children actually receive. More recently,~\citet{vong2024grounded} demonstrated that a contrastive image-text model akin to CLIP~\cite{radford2021learning} trained on headcam video and child-directed speech from the SAYCam dataset~\cite{sullivan2021saycam} can acquire meaningful word-referent mappings. However, their training procedure processes the data in an offline, i.i.d. fashion with up to 400 cycles over the full dataset. This setup contrasts with the single-pass, temporally structured nature of human learning---a recent critic argues that this data cycling invalidates any cognitively-meaningful conclusions drawn from these models~\cite{bowers2025successes}.

Is the challenge of single-pass word learning addressable? There are some promising results on vision-only learning.~\citet{zhuang2021unsupervised} established benchmarks for life-long visual learning on the SAYCam dataset~\cite{sullivan2021saycam} and evaluated a range of deep self-supervised algorithms, finding that contrastive methods that leverage negative sampling from sampling are most effective~\citep[e.g., SimCLR;][]{chen2020simple}. Other work offers ways of alleviating the temporal redundancy issue through more sophisticated buffers~\citep{purushwalkam2022challenges}. Most related to our method here,~\citet{yang2025memory} proposed Memory Storyboard, a continual visual representation learning framework that groups past frames into temporal segments and employs a temporal contrastive loss and a two-tier memory hierarchy to achieve higher performance on downstream vision tasks. 
While all of these studies focus exclusively on visual representations and do not address the language grounding aspect, leveraging the abundant visual information from egocentric videos could, in principle, make verbal learning also more efficient.

\begin{figure}[!t]
  \centering
  \includegraphics[width=0.98\textwidth]{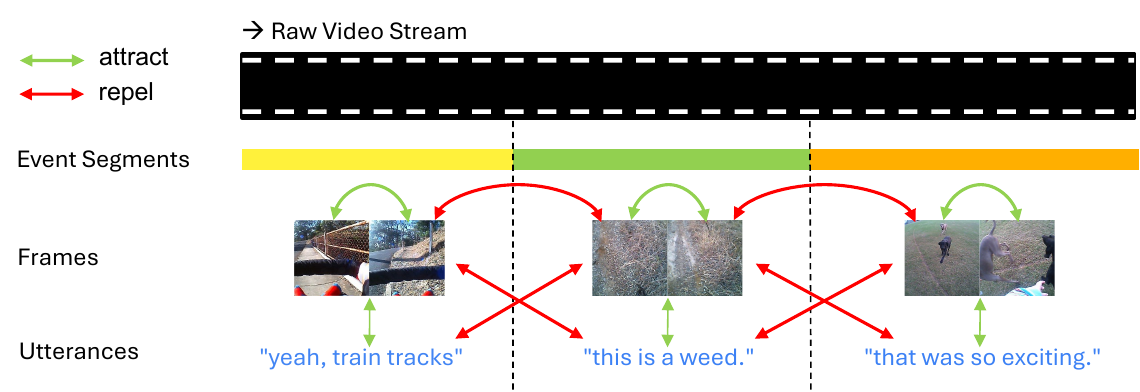}
  \caption{Overview of BabyCL. The incoming SAYCam stream is partitioned into event segments of roughly three minutes using hierarchical visual clustering, with boundaries adjusted to utterance timestamps. Embeddings of frames within the same event segment are brought together while those from different event segments are pushed apart ($\mathcal{L}_S$). Matching pairs of utterances and frames are brought together while mismatching ones are pushed apart ($\mathcal{L}_C$).}
  \label{fig:BabyCL_overview}
\end{figure}

Here, we bridge these lines of research: we present a continual 
visual and verbal learning framework that jointly learns visual representations and image-text associations, trained from scratch on child egocentric video and child-directed speech. Our proposed framework processes the data as a single chronological stream, consistent with how a young child lived these experiences. We use a modest-sized replay buffer to mitigate temporal redundancy and stabilize training. Our approach combines an image-only contrastive objective with an image-text contrastive objective, enabling the model to simultaneously learn a good visual representation and ground linguistic input. We show that meaningful image-text representations for word-referent mapping can emerge under these more cognitively realistic training settings.

\section{Related Work}
\paragraph{Developmentally inspired learning.} Several recent works have trained deep learning models on child-perspective sensory data, predominantly the SAYCam dataset~\citep{sullivan2021saycam}.~\citet{orhan2020self} showed the emergence of high-level visual category representations in self-supervised visual models trained on the temporal classification objective.~\citet{sheybani2023curriculum} demonstrated that the temporal ordering of infant experience provides a useful curriculum signal for visual representation learning.~\citet{yang2025memory} further showed that with a memory buffer significantly smaller in size than the total training frames, self-supervised learning models can achieve competitive performance to offline training in a single-pass, streaming setting. These results suggest that self-supervised learning methods are highly effective for learning good visual representations from child-perspective visual data, but they remain limited to the visual modality.

Extending to grounded language learning,~\citet{vong2024grounded} trained a contrastive vision-language model (CVCL) on egocentric video frames paired with transcribed child-directed speech from a single child in SAYCam, showing that the model acquires word-referent mappings and generalizes to novel referents.~\citet{vong2025robustness} established the robustness of these findings across all three children and multiple model architectures. Concurrently, BabyVLM~\citep{wang2025babyvlm, wang2025babyvlmv2} explored generative vision-language models trained on SAYCam data augmented with synthetic child-directed captions and introduced developmentally motivated evaluation benchmarks. These results provide empirical evidence that deep learning models can acquire word meanings from weakly supervised, naturally co-occurring vision-language pairs in naturalistic, continuous video streams. However, a key limitation is that these models were trained offline with shuffled data over multiple epochs, in contrast to how children experience their environment.~\citet{bowers2025successes} argues that such discrepancies between deep learning model training and child learning undermine any cognitively-relevant conclusions drawn from these models. Our work directly addresses this criticism by learning grounded multimodal representations in a single-pass, temporally ordered stream, demonstrating that the offline training assumption is not necessary for successful word-referent acquisition.

\paragraph{Continual representation learning.}
Continual Representation Learning~\citep{rao2019continual, fini2022self, gomez2022continually, zhang2024integrating} addresses the problem of learning good representations from unlabeled, non-stationary data streams.
\citet{purushwalkam2022challenges} identified temporal correlation and distribution shift as key challenges for continuous self-supervised learning and proposed minimum-redundancy (MinRed) replay buffers to mitigate these issues.~\citet{madaan2021representational} demonstrated that unsupervised representations are more robust to catastrophic forgetting than supervised ones and mixing current and past instances further alleviates forgetting. Most directly related to our setting,~\citet{yang2025memory} proposed Memory Storyboard, a framework for streaming self-supervised learning from long-form egocentric video, including the SAYCam dataset. Their approach uses temporal segmentation to organize a two-tier memory hierarchy, grouping recent frames into event-like segments for more effective replay.
Recent work has also begun to address continual learning for vision-language models~\cite{liu2025continual, liu2025cclip}. However, these works focus on adapting pretrained models to shifting task distributions, rather than learning multimodal representations from scratch in a developmentally motivated streaming setting. Our work differs in that we effectively combine a visual contrastive loss with an image-text contrastive loss to learn a joint image-text representation through child-centric data in a streaming setting.

\section{Method}
\label{sec:method}
We present \textbf{BabyCL}, a framework for continual grounded language learning from a continuous video stream paired with aligned text. The pipeline is illustrated in Figure~\ref{fig:BabyCL_overview} and simultaneously perform visual and verbal learning under a shared backbone: 1) a streaming visual representation learning from~\citet{yang2025memory}, which uses hierarchical temporal segmentation and a two-tier replay buffer, and 2) the CVCL image-text contrastive objective of~\citet{vong2024grounded}. At each step, BabyCL is jointly optimized with three contrastive losses using samples drawn from two parallel replay buffers. 

\paragraph{Temporal segmentation.} Following~\citet{yang2025memory}, we partition the incoming stream into event segments of roughly three minutes, obtained by hierarchical clustering of frame embeddings. Concretely, given the frame sequence $\mathcal{V}=(x_1,\dots,x_L)$ and an embedding network $f_\theta$, the algorithm partitions the stream into $n$ contiguous segments by selecting change points that maximize the average within-segment cosine similarity of frame embeddings. The number of segments is set to $n=\lfloor L/T \rfloor$, where $T$ is the average segment length and is treated as a hyperparameter; we use $T$ corresponding to three minutes at $5$ fps. The maximization is approximated in a greedy manner. To accommodate the online streaming learning setting, we apply temporal segmentation in the short-term memory. We further adapt this temporal segmentation algorithm to our multimodal setting by constraining the change points to fall outside any utterance interval, enforcing that every utterance from the transcript lies within a single event segment.
 
\paragraph{Memory replay.} To mitigate temporal redundancy and catastrophic forgetting in the single-pass stream, BabyCL maintains two replay buffers. The visual buffer stores past event segments for frame-level self-supervision. The multimodal buffer stores utterance-frame pairs; following~\citet{vong2024grounded}, a frame is randomly sampled from within each utterance's timestamp window. Each buffer is split into a short-term FIFO tier and a long-term reservoir-sampled tier, as in~\citet{yang2025memory}. At each step, we form a training batch by sampling equally from the two buffers, and within each buffer 25\% from the short tier and 75\% from the long. For the experiments reported here, the short and long tiers hold by default 10\% and 50\% of the total stream, respectively (which are hyperparameters we vary in Seciton~\ref{sec_buffer_size}) .

\paragraph{Resampling at insertion.} Because SAYCam is traversed in a single chronological pass, the total number of gradient steps is bounded by the length of the stream. To ensure that training converges within this budget, each time the newly finalized event segments are inserted into the visual buffer, we perform $k$ (8 by default) additional forward-backward passes on batches that mix the newly inserted material with replayed content. The chronological order in which new segments first enter the stream is preserved; resampling only affects how often past content is revisited during training.

\paragraph{Training objectives.} All samples are encoded by a shared ResNeXt-50 visual backbone (matching CVCL~\cite{vong2024grounded}), with modality-specific projection heads. BabyCL is trained with three losses applied jointly at every step:
\begin{itemize}
\item \textbf{Instance loss $\mathcal{L}_I$}, a SimCLR objective~\cite{chen2020simple} on augmented views of frames sampled from the visual buffer.
\item \textbf{Temporal loss $\mathcal{L}_S$}, a supervised classification loss over event segments~\citep{orhan2020self, yang2025memory}.
\item \textbf{Cross-modal loss $\mathcal{L}_C$}, a symmetric InfoNCE loss on utterance-frame pairs drawn from the multimodal buffer, matching CVCL~\cite{vong2024grounded}.
\end{itemize}
The three losses are combined with equal per-loss weight, giving the visual branch twice the aggregate weight of the multimodal branch:
\begin{equation}
\mathcal{L} = 0.5\,(\mathcal{L}_I + \mathcal{L}_S) + 0.5\,\mathcal{L}_C.
\end{equation}
All parameters, including the shared backbone and the modality-specific heads, are updated jointly in a single backward pass. Because the two visual losses and the cross-modal loss share a backbone, $\mathcal{L}_I$ and $\mathcal{L}_S$ shape a visual manifold reflecting the temporal structure of the child's experience, while $\mathcal{L}_C$ grounds that manifold in language.

\section{Experiments}

\subsection{Experiment setup}
We train BabyCL on the child S subset of the SAYCam~\citep{sullivan2021saycam} dataset, which contains 221 hours of video data collected from a head-mounted camera on the child from age 6-32 months, decoded at 5 fps. For multimodal supervision, we used the transcribed child-directed speech data curated by~\citet{vong2024grounded}, which includes 37,486 child-directed utterances paired with 600,285 image frames decoded at 5 fps. We evaluate BabyCL on Labeled-S using the 4AFC protocol of~\citet{vong2024grounded}, with controlled comparisons against CVCL-style baselines and an ablation of the replay buffer design.

We report accuracy on the four-alternative forced-choice (4AFC) task on Labeled-S~\citep{orhan2020self}, following~\citep{vong2024grounded, vong2025robustness}, shown in Figure~\ref{fig:4afc}. Each trial presents one target category and three foils; the model scores four candidate images against the target category name and must identify the matching one. Accuracy is the fraction of trials in which the target receives the highest score.

\providecommand{\FourAFCfigFrac}{0.46}
\providecommand{\FourAFCtabFrac}{0.52}
\providecommand{\FourAFCfigVspace}{0.5em}
\providecommand{\FourAFCtabVspace}{2.4em}

\begin{figure}[!t]
  \centering
  \begin{minipage}[t]{\FourAFCfigFrac\linewidth}\vspace{0pt}
    \centering
    \vspace{\FourAFCfigVspace}
    \includegraphics[width=0.8\linewidth]{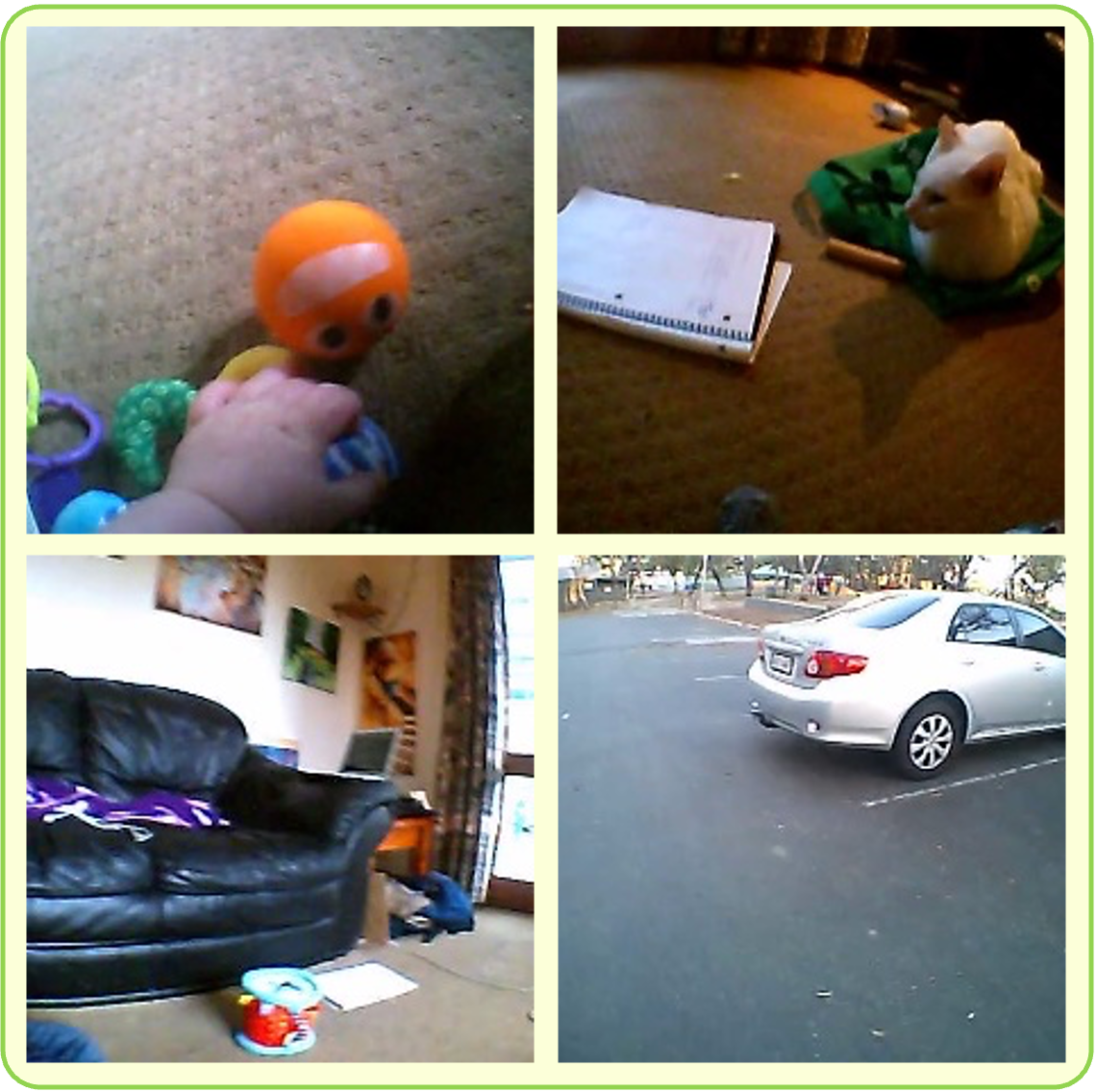}
    \caption{\textbf{Example of the 4AFC task.} Given a target category (``ball'') and three foil categories (``cat'', ``couch'', and ``car''), the model must select the image corresponding to the target category from four candidate images.}
    \label{fig:4afc}
  \end{minipage}
  \hfill
  \begin{minipage}[t]{\FourAFCtabFrac\linewidth}\vspace{0pt}
    \centering
    \small
    \vspace{\FourAFCtabVspace}
    \captionof{table}{Labeled-S 4AFC image-mode accuracy. Reported values show mean accuracy with approximate 95\% confidence intervals across seeds. Offline CVCL serves as an upper bound (i.i.d., 400 epochs); one-pass CVCL represents naive streaming. CL-CVCL matches BabyCL's gradient budget at $k=8$. BabyCL ($k\times$) denotes $k$ forward passes per video-buffer insertion.}
    \label{tab:streaming_cvcl}
    \vspace{0.25\baselineskip}
    \setlength{\tabcolsep}{6pt}
    \begin{tabular}[t]{lc}
      \toprule
      \tableheader Method & \tableheader 4AFC (\%) \\
      \midrule
      \rowwhite Offline CVCL (i.i.d.) & 57.31 $\pm$ 1.13 \\
      \midrule
      \rowwhite One-pass CVCL (streaming) & 27.52 $\pm$ 3.12 \\
      \rowwhite CL-CVCL (compute-matched) & 36.41 $\pm$ 2.31 \\
      \rowwhite BabyCL ($k=4$) & \secondbest{41.08 $\pm$ 1.79} \\
      \rowblue BabyCL ($k=8$) & \best{43.38 $\pm$ 1.47} \\
      \bottomrule
    \end{tabular}
  \end{minipage}
\end{figure}

\subsection{Main results}
We compare BabyCL to three CVCL-style baselines that isolate different explanations for streaming performance. \textbf{Offline CVCL}~\cite{vong2024grounded} trains on i.i.d.\ shuffled data for 400 epochs and serves as an upper bound for what the CVCL objective can achieve when temporal structure is removed. \textbf{One-pass CVCL} applies the CVCL objective to minibatches arriving in temporal order with no replay, representing naive streaming. \textbf{CL-CVCL} also streams the data in temporal order but introduces replay to match BabyCL's gradient budget at 
$k=8$ (approximately $6\times$ revisitation per item); it thus isolates the effect of BabyCL's continual-learning machinery from pure optimization count.

Table~\ref{tab:streaming_cvcl} reports 4AFC accuracy. One-pass CVCL collapses to 25.68\%, confirming that a training recipe designed for i.i.d. data does not transfer to streaming. CL-CVCL, which matches BabyCL's update budget but omits its continual-learning structure, reaches only 35.32\%, well below BabyCL at 44.46\% ($k=8$). Because CL-CVCL is matched in optimization work, the gap cannot be explained by gradient count alone; it reflects the benefit of organizing the stream through event segmentation and of combining visual self-supervision with cross-modal grounding. Additional insertion-time resampling (from $k=4$ to $k=8$) yields a further 2.6-point gain, indicating that the continual-structure benefits compound with more thorough revisitation of newly inserted content. BabyCL substantially closes the gap to offline CVCL (57.45\%) but does not match it; we view this remaining gap as an honest reminder that i.i.d.\ training with hundreds of epochs remains a strong reference point, and narrowing it further under realistic training will be an enduring research direction.

\subsection{Ablation experiments and other training factors} \label{sec_buffer_size}

\paragraph{Balancing Visual and Language-Grounded Supervision}
BabyCL is trained from two coupled sources of supervision: dense unlabeled video and sparse utterance-aligned image-text pairs. We investigate how different proportions of language-grounded examples in each batch affect training performance, and summarize the outcomes in Table~\ref{tab:mm_frac_ablation}. Specifically, we vary the fraction of examples drawn from the multimodal buffer in each batch among $25\%$, $50\%$, and $75\%$, with the remainder drawn from the visual buffer. All other hyperparameters remain the same as in the default setting; each setting uses three random seeds.

\providecommand{\MmPolicyLeftFrac}{0.52}
\providecommand{\MmPolicyRightFrac}{0.44}

\begin{table}[t]
  \centering
  \begin{minipage}[t]{\MmPolicyLeftFrac\linewidth}\vspace{0pt}
    \centering
    \small
    \captionof{table}{\textbf{Balancing visual and language-grounded supervision.} Labeled-S 4AFC when multimodal (MM) pairs occupy $25\%$, $50\%$, or $75\%$ of each minibatch ($n{=}3$ seeds). \emph{Best} is the largest peak accuracy observed across seeds during training. \emph{Final mean $\pm$ CI} is the mean end-of-training accuracy with an approximate $95\%$ confidence interval.}
    \label{tab:mm_frac_ablation}
    \vspace{0.25\baselineskip}
    \setlength{\tabcolsep}{5pt}
    \begin{tabular}[t]{lcc}
      \toprule
      \tableheader MM frac. & \tableheader Best (\%) & \tableheader Final mean $\pm$ CI (\%) \\
      \midrule
      \rowwhite $25\%$ & \secondbest{45.00} & \secondbest{43.38 $\pm$ 1.47} \\
      \rowblue $50\%$ & \best{45.35} & \best{43.43 $\pm$ 1.36} \\
      \rowwhite $75\%$ & 44.43 & 42.67 $\pm$ 1.29 \\
      \bottomrule
    \end{tabular}
  \end{minipage}
  \hfill
  \begin{minipage}[t]{\MmPolicyRightFrac\linewidth}\vspace{0pt}
    \centering
    \small
    \captionof{table}{Effect of replay buffer policy on Labeled-S 4AFC accuracy. The same policy is applied to both buffers. Each non-default row is the better of two random seeds.}
    \label{tab:buffer_ablation}
    \vspace{0.25\baselineskip}
    \setlength{\tabcolsep}{4pt}
    \begin{tabular}[t]{lc}
      \toprule
      \tableheader Policy & \tableheader 4AFC (\%) \\
      \midrule
      \rowwhite FIFO + random & 43.90 \\
      \rowwhite Single-tier reservoir & 44.00 \\
      \rowwhite FIFO + FIFO & \secondbest{44.31} \\
      \rowblue FIFO + reservoir (default) & \best{45.00} \\
      \bottomrule
    \end{tabular}
  \end{minipage}
\end{table}

The balanced $50\%$ mixture attains both the highest peak score and the strongest terminal mean among the three regimes: skewing toward fewer multimodal slots weakens grounding pressure, whereas overweighting multimodal batches dilutes coverage of the dense visual stream. Taken together, Table~\ref{tab:mm_frac_ablation} supports allocating comparable mass to temporally structured vision losses and utterance-aligned pairs rather than privileging either modality alone.

\paragraph{Robustness to replay buffer design} We investigate how sensitive BabyCL is to the specific buffer management policy, keeping all other training choices fixed. The same policy is applied to both the video and the multimodal buffers, with the 10\%/50\% short/long capacity split used in the main runs. We denote a two-tier policy by its \emph{short-tier + long-tier} eviction rules and compare the default configuration, \emph{FIFO + reservoir}, against three alternatives: \emph{FIFO + FIFO}, which replaces the long-tier reservoir with FIFO eviction; \emph{FIFO + random}, which evicts a uniformly chosen long-tier resident on overflow; and a \emph{single-tier reservoir} that merges both tiers into a flat reservoir of equal total capacity, removing the dedicated recent-item channel. Each alternative is run with two random seeds, and we report the better run.

As Table~\ref{tab:buffer_ablation} shows, all four policies fall within roughly half a percentage point of each other (43.90\%--44.46\%). BabyCL is therefore robust to the specific eviction rule: once a two-tier layout with a dedicated channel for recent items is in place, the choice of long-tier policy matters comparatively little.

\paragraph{Size of the visual buffer}
The default visual replay buffer in BabyCL is relatively large: the short-term and long-term tiers store examples equivalent to $10\%$ and $50\%$ of the full video stream, respectively. We acknowledge that this memory budget is substantial for long streams, and therefore evaluate whether BabyCL remains effective under smaller visual replay buffers. Specifically, we reduce the short-term/long-term buffer capacities to $5\%/25\%$ and $2\%/10\%$ of the stream while keeping the rest of the training recipe fixed. Results are shown in Table~\ref{tab:visual_buffer_ablation}.

\begin{table}[t]
  \centering
  \small
  \caption{\textbf{Effect of visual buffer size.} We vary the short-term and long-term visual buffer capacities as a percentage of the total video stream. Performance decreases as the buffer size is reduced, but BabyCL remains above the CL-CVCL baseline even with substantially smaller buffers.}
  \label{tab:visual_buffer_ablation}
  \setlength{\tabcolsep}{8pt}
  \begin{tabular}{lccc}
    \toprule
    \tableheader Method & \tableheader Short buffer & \tableheader Long buffer & \tableheader 4AFC (\%) \\
    \midrule
    \rowwhite CL-CVCL & - & - & 37.47 \\
    \rowwhite BabyCL & 2\% & 10\% & 38.46 \\
    \rowwhite BabyCL & 5\% & 25\% & \secondbest{41.57} \\
    \rowblue BabyCL & 10\% & 50\% & \best{45.00} \\
    \bottomrule
  \end{tabular}
\end{table}

We observe that reducing the visual buffer size leads to a clear drop in downstream performance, indicating that replay capacity is important for maintaining diverse visual experience from the stream. Nevertheless, BabyCL continues to outperform the CL-CVCL baseline even with substantially smaller buffers, such as a $2\%/10\%$ short-term/long-term buffer. This suggests that while larger visual buffers are beneficial, the advantage of BabyCL is not solely due to using a large memory budget.

\subsection{Linear probing without finetuning}
\label{sec:linear_probe}
Forced-choice alignment scores summarize how well vision and text co-index particular nouns; they do not, by themselves, describe how discriminable frozen visual features remain once language is reduced to a lightweight readout. To isolate purely visual structure we therefore train linear classifiers on top of fixed backbone embeddings while leaving convolutional weights untouched.

\providecommand{\ProbeVtwtLeftFrac}{0.50}
\providecommand{\ProbeVtwtRightFrac}{0.46}

\begin{table}[t]
  \centering
  \begin{minipage}[t]{\ProbeVtwtLeftFrac\linewidth}\vspace{0pt}
    \centering
    \small
    \captionof{table}{\textbf{Linear probing frozen visual embeddings.} Top-$1$ accuracy (\%) on infant-centric Labeled-S categories ($24$-way stratified split) and ImageNet Mini ($100$-way train/val). Checkpoints match the strongest Labeled-S 4AFC validation selection used elsewhere.}
    \label{tab:linear_probe}
    \vspace{0.25\baselineskip}
    \setlength{\tabcolsep}{5pt}
    \begin{tabular}[t]{lcc}
      \toprule
      \tableheader Method & \tableheader Labeled-S & \tableheader Mini-ImageNet \\
      \midrule
      \rowwhite Offline CVCL & \best{60.17} & \best{37.22} \\
      \midrule
      \rowwhite One-pass CVCL & 18.36 & 2.92 \\
      \rowwhite CL-CVCL & 29.32 & 7.50 \\
      \rowwhite BabyCL & \secondbest{45.69} & \secondbest{25.40} \\
      \bottomrule
    \end{tabular}
  \end{minipage}
  \hfill
  \begin{minipage}[t]{\ProbeVtwtRightFrac\linewidth}\vspace{0pt}
    \centering
    \small
    \captionof{table}{\textbf{VTWT zero-shot binary retrieval.} Accuracy (\%) when cosine similarity ranks the image with the correct phrase above the foil. Checkpoints match the Labeled-S 4AFC validation criterion used elsewhere; VTWT inference performs no additional training.}
    \label{tab:vtwt}
    \vspace{0.25\baselineskip}
    \setlength{\tabcolsep}{8pt}
    \begin{tabular}[t]{lc}
      \toprule
      \tableheader Method & \tableheader VTWT (\%) \\
      \midrule
      \rowwhite Offline CVCL & \best{63.08} \\
      \midrule
      \rowwhite One-pass CVCL & 52.22 \\
      \rowwhite CL-CVCL & 55.43 \\
      \rowwhite BabyCL & \secondbest{61.01} \\
      \bottomrule
    \end{tabular}
  \end{minipage}
\end{table}

Embeddings are extracted at $512$ dimensions after the frozen encoder, $\ell_2$-normalized, and passed to a multinomial logistic regression trained with mild $\ell_2$ regularization. The SAYCam-side probe covers twenty-four infant-centric object categories with an 80/20 stratified train/test partition; ImageNet Mini supplies a complementary hundred-way recognition benchmark fit on the canonical training split and scored on validation frames. Input preprocessing mirrors each model family at train time (CLIP channel statistics for Offline CVCL and the streaming CVCL baselines; ImageNet statistics for BabyCL). Model checkpoints correspond to the strongest Labeled-S 4AFC validation performance, identical to the criterion used in preceding sections.

Table~\ref{tab:linear_probe} places Offline CVCL ahead on both probes ($60.17\%$ / $37.22\%$), with BabyCL second ($45.69\%$ / $25.40\%$), matched-budget CL-CVCL at $29.32\%$ / $7.50\%$, and one-pass streaming lowest ($18.36\%$ / $2.92\%$). Absolute scores compress sharply away from SAYCam-aligned categories toward ImageNet viewpoints, yet the streaming-family ordering is unchanged. BabyCL gains $+16.37$ percentage points over matched replay on Labeled-S and $+17.90$ on ImageNet Mini, suggesting that continual segmentation with insertion-synchronized revisitation sharpens frozen geometry rather than merely lifting alignment-centric benchmarks.

\subsection{Visual Two-Word Test (VTWT) and Baby Winoground}
\label{sec:babyvlm}

Visual Two-Word Test (VTWT) and Baby Winoground are two zero-shot, in-domain evaluation benchmarks proposed by~\citet{wang2025babyvlm} for multimodal models trained on SAYCam.

\paragraph{Visual Two-Word Test (VTWT).} VTWT probes compositionality by asking the model to match SAYCam frames with appropriate two-word phrases (e.g., ``wash cup'' vs. ``fill cup''). For VTWT evaluation, we use the same image preprocessing pipeline as the linear-probe evaluations. Text spans are embedded by averaging word vectors prior to projection; vision and language descriptors are $\ell_2$-normalized before cosine scoring, and an item counts correct when the faithful caption aligns more tightly with the frame than the foil does.

Table~\ref{tab:vtwt} reports accuracy above chance ($50\%$) for every system but with a wide spread: Offline CVCL leads at $63.08\%$, BabyCL reaches $61.01\%$, matched-budget CL-CVCL attains $55.43\%$, and one-pass CVCL $52.22\%$. The gain from adding replay budget without BabyCL's continual structure ($52.22\%$ to $55.43\%$) is modest next to BabyCL's margin over matched replay ($+5.58$ percentage points), consistent with the view that replay alone does not recover fine phrase distinctions unless paired with temporally coherent segmentation as in BabyCL.

\paragraph{Baby Winoground.} Baby Winoground couples two simultaneously observed scenes with the same contrasting phrase pair such that each image warrants exactly one wording under the intended semantics. The benchmark evaluates whether a model can preserve correct image--text associations under both the original SAYCam context and a synthetically perturbed negative context. A successful prediction therefore requires dual consistency: the original scene must prefer its faithful caption over the foil, while the edited companion scene must symmetrically favor the alternate phrase.

\begin{table}[t]
  \centering
  \caption{\textbf{Baby Winoground (zero-shot).} Accuracy (\%). \emph{Overall} requires both association checks jointly. \emph{Positive Ctx} measures correctness under the original SAYCam context, while \emph{Negative Ctx} evaluates robustness under the synthetic negative context.}
  \label{tab:baby_winoground}
  \small
  \setlength{\tabcolsep}{6pt}
  \begin{tabular}{lccc}
    \toprule
    \tableheader Method & \tableheader Overall & \tableheader Positive Ctx & \tableheader Negative Ctx \\
    \midrule
    \rowwhite Offline CVCL & \secondbest{11.51} & \best{63.01} & 41.92 \\
    \rowwhite One-pass CVCL & 3.29 & 48.22 & \best{48.77} \\
    \rowwhite CL-CVCL & 8.22 & 55.34 & \secondbest{45.21} \\
    \rowblue BabyCL & \best{12.88} & \secondbest{59.73} & 41.37 \\
    \bottomrule
  \end{tabular}
\end{table}

Embeddings follow the VTWT recipe ($512$ dimensions after projection, $\ell_2$-normalized cosines). Pairwise similarities induce two complementary ranking constraints: one under the original visual context and the other under the synthetic negative context. Overall accuracy counts only trials where both conditions are simultaneously satisfied.

Table~\ref{tab:baby_winoground} summarizes the outcome. Offline CVCL achieves the strongest Positive Context score ($63.01\%$) but drops substantially under the Negative Context condition ($41.92\%$), suggesting weaker robustness once the visual semantics are perturbed. BabyCL is slightly lower on the original-context evaluation ($59.73\%$) yet attains the best Overall score ($12.88\%$), indicating more stable alignment across both contexts simultaneously. One-pass CVCL remains near the $50\%$ random baseline on both context-specific metrics ($48.22\%$ / $48.77\%$), while Overall accuracy collapses to $3.29\%$, implying that the model struggles to maintain coherent pairwise associations across the two contexts jointly. Matched-budget CL-CVCL lies between these extremes, improving over one-pass training but remaining below both Offline CVCL and BabyCL.

Overall, the results suggest that offline training favors alignment under the original SAYCam distribution but generalizes less reliably to synthetically perturbed contexts. BabyCL sacrifices a small amount of in-context accuracy yet produces representations that remain more consistent under context shifts, leading to stronger joint performance on the full Baby Winoground criterion.

\subsection{Nearest-neighbor segment retrieval}
\label{sec:nearest_neighbor}
The quantitative probes constrain hypotheses within scripted contrasts; we add a lightweight qualitative check that cosine neighborhoods of frozen embeddings reflect recurring infant-centric scenes rather than spurious similarity along individual recordings. Contiguous SAYCam intervals come from hierarchical temporal clustering~\citep{yang2025memory}; for each interval we average $512$-dimensional backbone descriptors over sampled frames, apply $\ell_2$ normalization, and rank candidates by cosine similarity while excluding every interval from the query recording itself.

\begin{figure}[t]
  \centering
  \includegraphics[width=\linewidth]{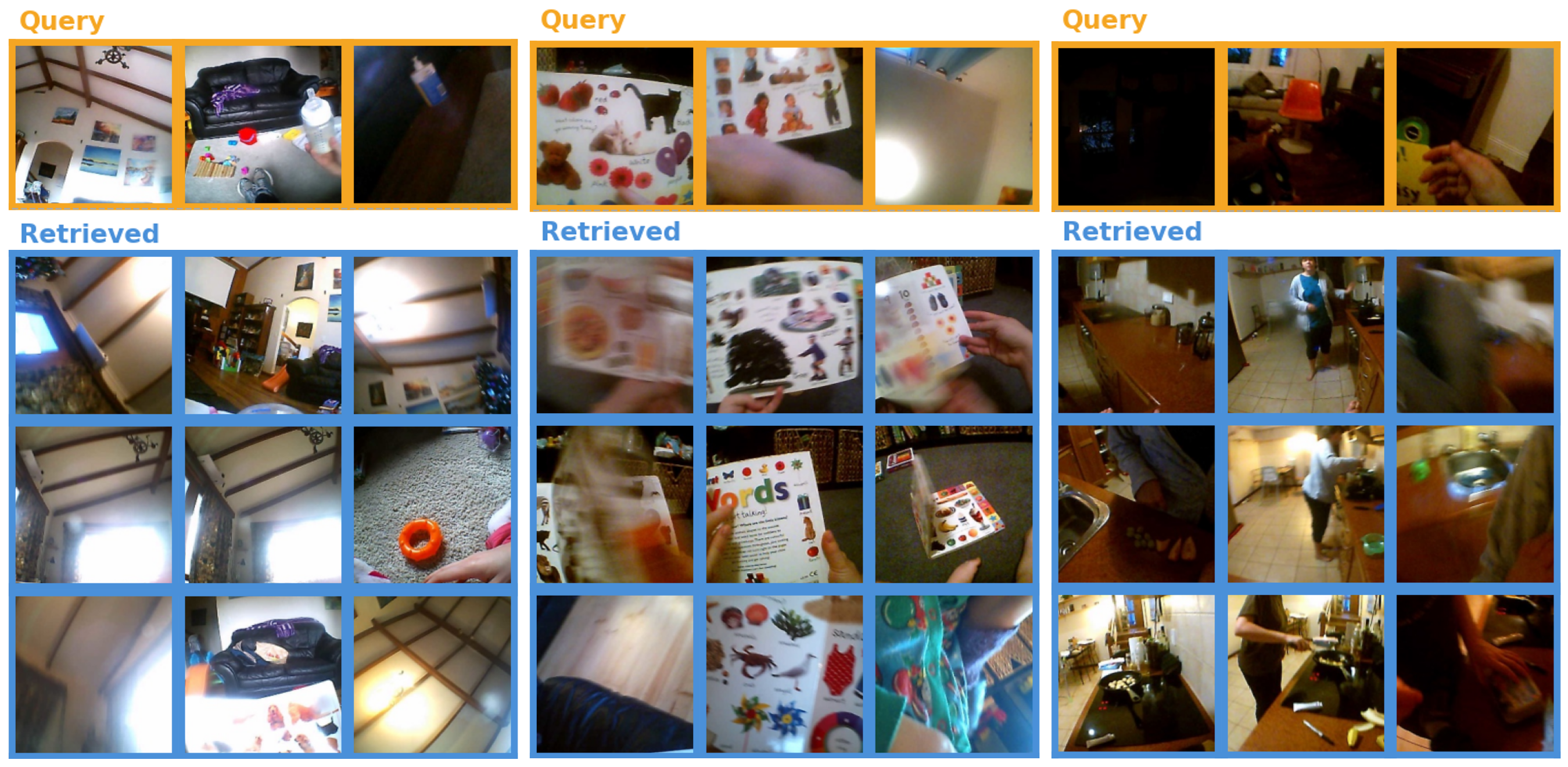}
  \caption{\textbf{Cross-video nearest neighbors (BabyCL).} Each major column corresponds to one query segment. The top row (yellow) shows representative frames from the query segment, while the rows below (blue) show representative frames from retrieved segments from other sessions.}
  \label{fig:nearest_neighbor}
\end{figure}

Figure~\ref{fig:nearest_neighbor} retrieves patios with play mats, picture-book interactions, and indoor kitchens across disjoint SAYCam captures---patterns aligned with the event-level supervision used in continual training rather than mere temporal continuity along one trajectory.

\subsection{Grad-CAM++ localization under noun queries}
\label{sec:gradcam}
Nearest-neighbor retrieval summarizes coarse neighborhoods; Grad-CAM++~\citep{chattopadhay2018grad} localizes where cosine alignment peaks inside each frame. For several Labeled-S nouns we retain infant-view exemplars that BabyCL scores most confidently against the embedded category token, backpropagate that cosine through the deepest convolutional maps of the frozen backbone, and overlay upsampled attributions at full resolution.

\begin{figure}[t]
  \centering
  \includegraphics[width=\linewidth]{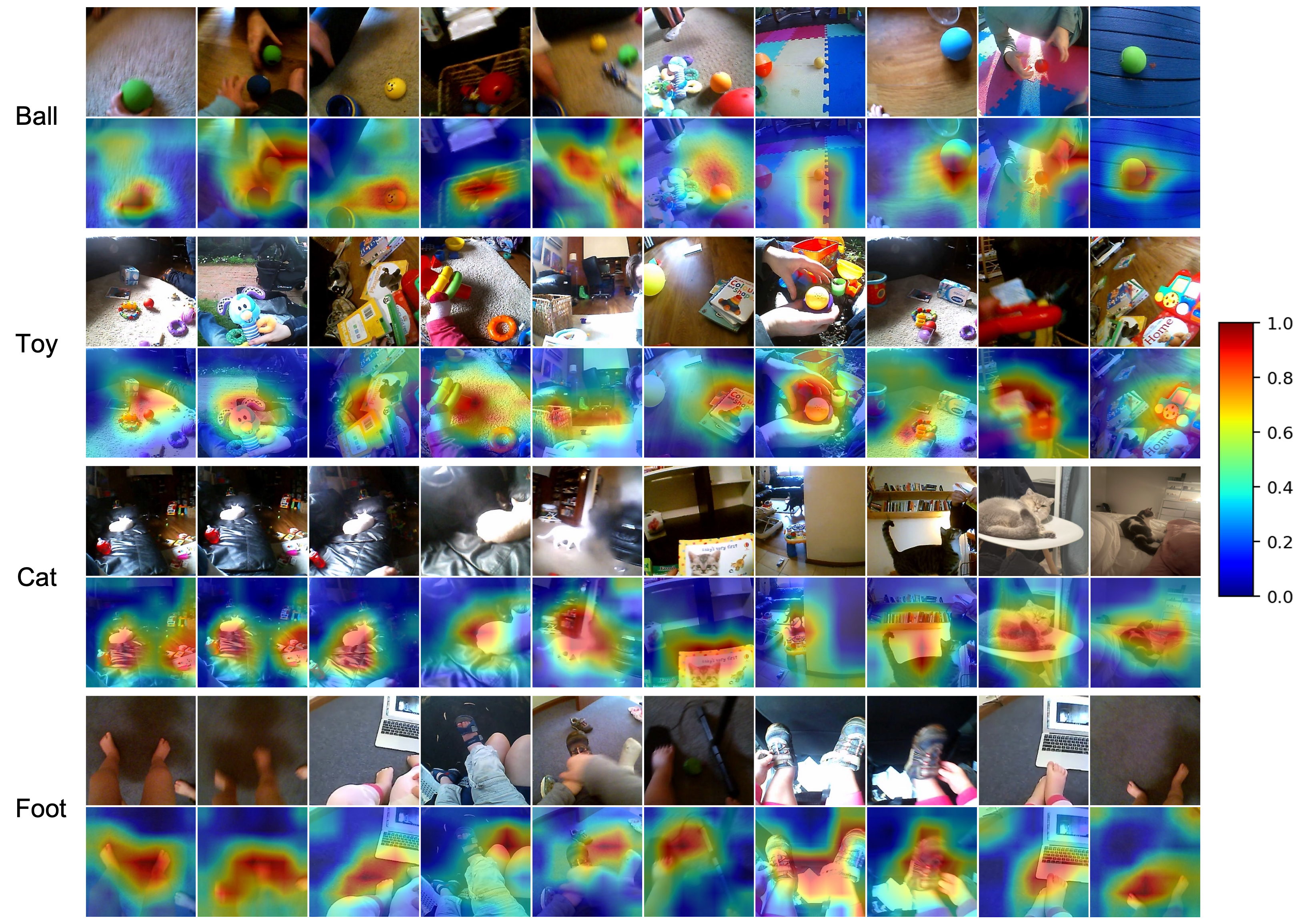}
  \caption{\textbf{Grad-CAM++ with noun queries (BabyCL).} Four Labeled-S categories; original frames appear above heatmap overlays for each.}
  \label{fig:gradcam}
\end{figure}

Figure~\ref{fig:gradcam} concentrates activation on balls, picture books and small toys, cats, and feet or shoes while leaving much of the cluttered domestic background cool---spatial sensitivity consistent with the cross-modal objective $\mathcal{L}_C$ encouraging noun-specific coupling rather than undifferentiated scene summaries.

\section{Discussion and Conclusion}
We introduced BabyCL, a framework for grounded word learning from a single chronological pass over child-perspective video and speech. By combining streaming visual self-supervision with a cross-modal contrastive objective under a dual replay buffer, BabyCL acquires meaningful word--referent mappings without the offline, i.i.d., multi-epoch training that characterizes prior work in this space. Under a matched optimization budget, BabyCL outperforms a streaming CVCL baseline by a wide margin on Labeled-S 4AFC, and is robust to the specific replay eviction policy.

These results show that grounded learning models can be trained in a continual streaming regime, which is much closer to a child's experience than traditional multi-epoch i.i.d. learning, directly addressing the concern raised by~\citet{bowers2025successes}.
Our findings suggest that this departure is not a necessary condition for word-referent acquisition: useful multimodal structure can emerge from data presented in the order a child actually encountered it.

\paragraph{Limitations and future work.} There remains a notable performance gap between BabyCL and offline CVCL; future works can aim to further close this gap. The replay buffers used here are large relative to the size of the video stream. Finally, additional evaluation benchmarks probing compositionality and generalization would give a more comprehensive picture.

\paragraph{Broader impact.} By moving grounded word learning closer to a single-pass, temporally structured regime, our work may support more realistic models of early learning and enable data-efficient training in long-form video settings. At the same time, any progress on learning from egocentric recordings raises privacy concerns; we rely on authorized-access datasets and advocate careful governance and restricted redistribution for sensitive data.

\newpage
\section*{Acknowledgements}
This research was supported by NYU-KAIST Award A25-0081-002, NSF BCS Award 2545541, Visko AI, Toyota Research Institute R2I program, a Google TPU Award, and the Institute of Information \& Communications Technology Planning Evaluation (IITP) under grant RS-2024-00469482, funded by the Ministry of Science and ICT (MSIT) of the Republic of Korea in connection with the Global AI Frontier Lab International Collaborative Research. The compute is supported by the NYU High Performance Computing resources, services, and staff expertise.

\bibliographystyle{apalike}
\bibliography{ref}

\newpage
\appendix

\section{Implementation details}
\label{sec:impl_details}
This appendix records concrete architectural and optimization details for BabyCL and the CVCL-family baselines.

\paragraph{Encoders and projection layers.}
All models use a DINO ResNeXt-50 vision backbone. We replace the classifier head with a single linear layer that maps the final ResNeXt block output (2048-d) to a 512-d embedding. The text side uses a learned 512-d word embedding table; an utterance embedding is the mean of its token embeddings (with out-of-vocabulary tokens mapped to the unknown-token index).

\paragraph{BabyCL: streaming visual branch.}
The visual branch uses a SimCLR-style objective with an MLP projector of sizes \([d, 2048, 128]\) (ReLU between layers), where \(d\) is the backbone feature dimension (2048 for ResNeXt-50). NT-Xent on paired augmented views uses temperature 0.2. The temporal classification component uses a second MLP of sizes \([d, 2048, C]\), where \(C\) is the number of temporal classes used by the TCL objective.

\paragraph{BabyCL: multimodal branch.}
The multimodal branch encodes images with the same vision backbone (and the 2048\(\rightarrow\)512 projection) and encodes text with the 512-d mean-pooled token embeddings. Image and text embeddings are \(\ell_2\)-normalized; we optimize symmetric image--text InfoNCE at temperature 0.07 (distinct from the visual NT-Xent temperature). Replay mixes short- and long-term buffers without semantic similarity filtering.

\paragraph{BabyCL: optimizer and schedule.}
We optimize with AdamW under automatic mixed precision, using the usual moment hyperparameters (\(\beta_1{=}0.9\), \(\beta_2{=}0.999\)). Four learning-rate tiers apply: \(10^{-3}\) for the shared ResNeXt convolutional stack, the embedding head that maps pooled features to the \(512\)-dimensional image embedding, and other decaying weights in that encoder (batch-normalization affine parameters and biases are trained at the same rate but omit weight decay), \(10^{-4}\) for the SimCLR and temporal-classification heads built on the \(2048\)-dimensional backbone output in the streaming branch, \(10^{-4}\) for multimodal modules outside that encoder, and \(10^{-4}\) for the small language-model stack excluding the shared token embedding table. Weight decay is \(10^{-4}\) wherever applied; scale and bias parameters in batch-normalization layers (and analogous no-decay terms in the encoder) use zero weight decay. After a linear warmup lasting 2\% of the scheduled optimizer steps (at least 100 steps), every tier follows the same cosine multiplier that decays to zero by the end of training. The main online runs use batch size 128.

\paragraph{BabyCL: data augmentation.}
For SSL we generate two views per frame using strong augmentation: random resized crop to \(224\times224\) (scale 0.2--1.0), random horizontal flip, color jitter (0.8/0.8/0.8/0.2 applied with 0.8 probability), random grayscale (0.2), Gaussian blur (applied with probability 1.0 on view 1 and 0.1 on view 2), and solarization on view 2 with probability 0.2. Inputs are normalized with ImageNet channel statistics.

\paragraph{BabyCL: online segmentation and replay composition.}
Online segmentation targets roughly 3-minute segments with embedding refresh every 60 frames. Each step trains on 128 frames in four 32-frame segments: three for streaming SSL/TCL (96 frames) and one for multimodal pairs (32 frames, one frame per aligned utterance chunk). The chronological stream advances once every eight optimization steps; intermediate steps resample from replay. Visual and multimodal replay both retain a short FIFO sized to 10\% of the stream and a long reservoir sized to 50\%; multimodal minibatches draw one quarter of chunks from the short tier and the rest from the long tier. The visual and multimodal losses are weighted equally in the total objective.

\paragraph{CVCL baselines (offline vs.\ streaming replay schedules).}
All variants share the same encoder stack and a cross-modal InfoNCE objective on paired frames and utterances. Offline CVCL trains with shuffled minibatches for multiple epochs (batch size 8). \emph{Streaming CVCL} (both One-pass and matched rows in our tables) mirrors BabyCL's continual scaffolding on chronological SAYCam: new clips advance into short FIFO first and spill into a capped long reservoir; minibatches always recombine short- and long-reservoir draws after unseen indices arrive (ratio one-to-three within each minibatch). The sole knob is rehearsal cadence after each buffer admission: \emph{one-pass} draws exactly one reservoir minibatch and trains once before more chronological indices are admitted; \emph{matched} draws six independent reservoir minibatches and trains six times before the next admission wave.

\paragraph{CVCL optimization.}
Offline CVCL uses AdamW (\(10^{-4}\) learning rate, weight decay \(0.1\)) with ReduceLROnPlateau driven by 4AFC validation accuracy. Streaming CVCL schedules share AdamW without that plateau loop during step-wise continual training and optimize multimodal InfoNCE only (language-model and auxiliary vision losses are disabled).

\section{Compute resources}
\label{sec:compute}
All BabyCL online runs reported here were executed on single NVIDIA L40S GPUs. Under the default training configuration (batch size 128, mixed precision, and the replay schedule described above), observed peak device memory was on the order of $20$\,GB.

\section{Licenses for existing assets}
\label{sec:asset_licenses}
The original papers for all the code, data, and models used in the main experiments of the paper are cited. The terms are properly respected. SAYCam-related video and derived annotations are accessed through Databrary under authorized-access and data-use agreements (not redistributed). Our implementation is provided in anonymized form as supplementary material and released under the MIT License. Pretrained vision backbones are obtained via the publicly released silicon-menagerie utilities and corresponding checkpoint distributions (MIT License). Baby Winoground and VTWT evaluation code/data labels follow the licensing of the BabyVLM evaluation repository: the core pipeline structure inherits the MIT License from lm-evaluation-harness, while the added multimodal tasks/models are released under Apache License 2.0. Core software dependencies include PyTorch and torchvision (BSD 3-Clause). Common scientific libraries used by our code are NumPy, SciPy, pandas, and scikit-learn (BSD 3-Clause), Pillow (HPND), and Matplotlib (PSF-based).

\end{document}